\documentclass[nohyperref]{article}

\usepackage[usenames,dvipsnames]{xcolor} 

\usepackage{arxiv}
\usepackage{microtype}
\usepackage{graphicx}
\usepackage{subfigure}
\usepackage{booktabs} 

\usepackage[colorlinks,linktoc=all]{hyperref} 

\usepackage{amsmath}
\usepackage{amssymb}
\usepackage{mathtools}
\usepackage{amsthm}

\usepackage[capitalize,noabbrev]{cleveref}

\theoremstyle{plain}

\theoremstyle{definition}

\theoremstyle{remark}

\usepackage[textsize=tiny]{todonotes}

\usepackage{natbib}
\usepackage{xspace}
\usepackage[all]{hypcap}
\hypersetup{citecolor=MidnightBlue}
\hypersetup{linkcolor=MidnightBlue}
\hypersetup{urlcolor=MidnightBlue}

\newcommand{\dgraph}{d_G} 
\newcommand{\activ}[1]{h^{(l)}} 
\newcommand{\readout}{r}

\newcommand{\defeq}{\stackrel{\text{\tiny def}}{=}}
\newcommand{\relu}{\mathrm{ReLU}}
\newcommand{\leakyrelu}{\mathrm{LeakyReLU}}
\newcommand{\elu}{\mathrm{ELU}}
\newcommand{\softmax}{\mathrm{softmax}}
\newcommand{\neigh}[1]{\cN_{#1}} 

\newcommand{\R}{\mathbb{R}}

\newcommand{\mathbold}[1]{\ensuremath{\boldsymbol{\mathbf{#1}}}}

\newcommand{\g}{\,|\,}

\newcommand{\nestedmathbold}[1]{{\mathbold{#1}}}

\newcommand{\mbh}{\nestedmathbold{h}}

\newcommand{\mby}{\nestedmathbold{y}}
\newcommand{\mbz}{\nestedmathbold{z}}

\newcommand{\mbH}{\nestedmathbold{H}}

\newcommand{\mbW}{\nestedmathbold{W}}
\newcommand{\mbX}{\nestedmathbold{X}}
\newcommand{\mbY}{\nestedmathbold{Y}}
\newcommand{\mbZ}{\nestedmathbold{Z}}

\newcommand{\mbalpha}{\nestedmathbold{\alpha}}

\newcommand{\mbOmega}{\nestedmathbold{\Omega}}

\newcommand{\cL}{\mathcal{L}}
\newcommand{\cN}{\mathcal{N}}

\newcommand{\esol}{ESOL\xspace}
\newcommand{\lipo}{Lipophilicity\xspace}
\newcommand{\enzymes}{Enzymes\xspace}
\newcommand{\cora}{Cora\xspace}
\newcommand{\citeseer}{Citeseer\xspace}
\newcommand{\pubmed}{Pubmed\xspace}
\title{Addressing Over-Smoothing in Graph Neural Networks via Deep Supervision}

\author{ {Pantelis Elinas} \\
	CSIRO's DATA61\\
	Sydney, AU \\
	\texttt{pantelis.elinas@data61.csiro.au} \\
	\And
	{Edwin V.~Bonilla} \\
	CSIRO's DATA61\\
	Sydney, AU \\
	\texttt{edwin.bonilla@data61.csiro.au} \\
}

\renewcommand{\shorttitle}{\textit{arXiv} Template}

\hypersetup{
pdftitle={Addressing Over-Smoothing in Graph Neural Networks via Deep Supervision},
pdfsubject={cs.LG, stat.ML},
pdfauthor={Pantelis Elinas, Edwin V.~Bonilla},
pdfkeywords={graph neural networks, deep supervision, representation learning},
}
\date{} 
\begin{document}
\maketitle
\begin{abstract}
Learning useful node and graph representations with graph neural networks (GNNs) is a challenging task. It is known that deep GNNs suffer from over-smoothing where, as the number of layers increases, node representations become nearly indistinguishable and model performance on the downstream task degrades significantly. 
To address this problem, we propose deeply-supervised GNNs (DSGNNs), i.e., GNNs enhanced with deep supervision where representations learned at all layers are used for training. We show empirically that DSGNNs are resilient to over-smoothing and can outperform competitive benchmarks on node and graph property prediction problems.

\end{abstract}
\section{Introduction}
\label{sec:introduction}
We live in a connected world and generate vast amounts of graph-structured or network data. Reasoning with graph-structured data has many important applications such as traffic speed prediction, product recommendation, and drug discovery~\citep{zhou2020graph, 10.1093/bib/bbab159}. Graph neural networks (GNNs), first introduced by~\citet{scarselli_gnn}, have emerged as the dominant solution for graph representation learning, which is the first step in building predictive models for graph-structured data. 

One of the most important applications of GNNs is that of \emph{node property prediction}, as in semi-supervised classification of papers (nodes) in a citation network \citep[see, e.g.,][]{kipf2016semi}. 
%
Another exciting and popular application of GNNs is that of \emph{graph property prediction}, as in, for example,  graph classification and regression. In this setting, we are given a set of graphs and corresponding labels, one for each graph, and the goal is to learn a mapping from the graph to its label. In both problems, node and graph property prediction, the labels can be binary, multi-class, multi-label, or continuous.  

Even though GNNs have been shown to be a powerful tool for graph representation learning, they are limited in depth, that is the number of GNN layers. Indeed, deep GNNs suffer from the problem of over-smoothing where, as the number of layers increases, the node representations become nearly indistinguishable and model performance on the downstream task deteriorates significantly. 

Previous work has analyzed and quantified the over-smoothing problem \citep{deep_gnn_meng20,Zhao2020PairNorm,chen2020measuring} as well as proposed methodologies to address it explicitly \citep{Li_Han_Wu_2018,Zhao2020PairNorm,jumping-knowledge-networks-xu18c}. 
Some of the most recent approaches have mainly focused on forcing diversity on latent node representations via normalization \citep[see, e.g,][]{zhou-et-al-neurips-2020,Zhao2020PairNorm}. However, while these approaches have tackled the over-smoothing problem in node-property prediction tasks with reasonable success, they have largely overlooked the graph-property prediction problem.   

In this paper we show that over-smoothing is also a critical problem in graph-property prediction and propose a different approach to overcome it. In particular, our method trains predictors using node/graph representations from all layers, each contributing to the loss function equally, therefore encouraging the GNN to learn discriminative features at all GNN depths. Inspired by the work of \citet{pmlr-v38-lee15a}, we name our approach deeply-supervised graph neural networks (DSGNNs). 

Compared to approaches such as those by \citet{deep_gnn_meng20}, our method only requires a small number of additional parameters that grow linearly (instead of quadratically) with the number of GNN layers. Furthermore, our approach can be combined with previously proposed methods such as  normalization~\citep{Zhao2020PairNorm} easily and we explore the effectiveness of this combination empirically. Finally, our approach is suitable for tackling \emph{both} graph and node property prediction problems.

In summary, our contributions are the following,
\begin{itemize}
    \item We propose the use of deep supervision for training GNNs, which encourages learning of discriminative features at all GNN layers. We refer to these types of methods as deeply-supervised graph neural networks (DSGNNs);
    \item DSGNNs can be used to tackle both node and graph-level property prediction tasks;
    \item DSGNNs are general and can be combined with any state-of-the-art GNN adding only a small number of additional parameters that grows linearly with the number of hidden layers and not the size of the graph;
    \item and we show that DSGNNs are resilient to the over-smoothing problem in deep networks and can outperform competing methods on challenging datasets.
\end{itemize}

\section{Related Work}
\label{sec:related-work}
GNNs have received a lot of attention over the last few years with several extensions and improvements on the original model of \citet{scarselli_gnn} including attention mechanisms \citep{velickovic2018graph} and scalability to large graphs \citep{hamilton2017inductive, klicpera2018predict, chiang2019cluster, zeng2019graphsaint}. While a comprehensive introduction to graph representation learning can be found in \citet{hamilton-grl-book-2020}, below we discuss previous work on graph property prediction and over-smoothing in GNNs focused on the node property prediction problem.

\subsection{Graph Property Prediction}
A common approach for graph property prediction is to first learn node representations using any of many existing GNNs~\citep{kipf2016semi, hamilton2017inductive, velickovic2018graph, xu2018powerful} and then aggregate the node representations to output a graph-level representation. Aggregation is performed using a readout (also known as pooling) function applied to the node representations output at the last GNN layer. A major issue is the readout function's ability to handle isomorphic graphs. It should be invariant to such permutations. Robustness for isomorphic graphs can be achieved via readout functions invariant to the node order such as the sum, mean or max.

Several more sophisticated readout functions have also been proposed. For example,~\citet{sag_pool} proposed weighted average readout using self-attention (SAGPool).~\citet{zhang2018end} proposed a pooling layer (SortPool) that sorts nodes based on their structural role in the graph; sorting the nodes makes them invariant to node order such that representations can be learnt using $1$D convolutional layers.


\Citet{graph-u-nets-gao19a} combine pooling with graph coarsening to train hierarchical graph neural networks. Similarly,~\citet{ying2018hierarchical} proposed differentiable pooling (DiffPool) where the pooling layer learns a soft assignment vector for each node to a cluster. Each cluster is represented by a single super-node and collectively all super-nodes represent a coarse version of the graph; representations for each super-node are learnt using a graph convolutional layer. These hierarchical methods coarsen the graph by reducing the graph's nodes at each convolutional layer down to a  single node whose representation is used as input to a classifier. 
\subsection{Over-smoothing in Node-Property Prediction}
\Citet{Li_Han_Wu_2018} focus on semi-supervised node classification in a setting with low label rates. They identify the over-smoothing problem as a consequence of the neighborhood aggregation step in GNNs; they show that the latter is equivalent to repeated application of Laplacian smoothing leading to over-smoothing. They propose a solution that increases the number of training examples using a random walk-based procedure to identify similar nodes. The expanded set of labeled examples is used to train a Graph Convolutional Network~\citep[GCN,][]{kipf2016semi}. The subset of nodes that the GCN model predicts most confidently are then added to the training set and the model is further fine-tuned; they refer to the latter process as self-training. This approach is not suitable for the graph property prediction setting where node-level labels are not available and the graphs are too small for this scheme to be effective.

\Citet{deep_gnn_meng20} propose Deep Adaptive Graph Neural Networks (DAGNNs) for training deep GNNs by separating feature transformation from propagation. DAGNN uses a Multi-layer Perceptron (MLP) for feature transformation and smoothing via powers of the adjacency matrix for propagation similarly to~\citet{klicpera2018combining} and~\citet{wu2019simplifying}. 
However, the cost of their propagation operation increases quadratically as a function of the number of nodes in the graph hence DAGNNs do not scale well to large graphs. Furthermore, DAGNN's ability to combine local and global neighborhood information is of limited use in the graph property prediction setting where the graphs are small and the distinction between local and global node neighborhoods is difficult to make.

\Citet{Zhao2020PairNorm} also analyze the over-smoothing problem and quantify it by measuring the row and column-wise similarity of learnt node representations. They introduce a normalization layer called PairNorm that during training forces these representations to remain distinct across node clusters. They show that generally, the normalization layer reduces the effects of over-smoothing for deep GNNs. 
To evaluate their approach, they identify the Missing Features (MF) setting such that when test node features are missing then GNNs with PairNorm substantially outperform GNNs without it. PairNorm is a general normalisation layer and it can be used with any graph GNN architecture including ours introduced in \cref{sec:ds_gnns}. It is applicable to both node and graph-level representation learning tasks.

\citet{zhou-et-al-neurips-2020} also adopt group normalisation ~\citep{wu2018group} in neural networks to the graph domain. They show their approach tackles over-smoothing better than PairNorm. Group normalisation is most suited to node classification tasks and requires clustering nodes into groups posed as part of the learning problem. The number of groups is difficult to determine and must be tuned as a hyper-parameter.

Finally, in an approach closely related to ours, \citet{jumping-knowledge-networks-xu18c} propose jumping knowledge networks (JKNets) that make use of jump connections wiring the outputs of the hidden graph convolutional layers directly to the output layer. These vectors are combined and used as input to a classification or regression layer. JKNets combine learnt node representations aggregated over different size neighborhoods in order to alleviate the problem of over-smoothing. We propose a different approach that, instead of combining hidden representations across layers, introduces a classification or regression layer attached to the output of each hidden graph convolutional layer.

\section{Graph Neural Networks}
\label{sec:graph_convolutional_networks}

Let a graph be represented as the tuple $G=(V, E)$ where $V$ is the set of nodes and $E$ the set of edges. The graph has $|V| = N$ nodes. We assume that each node $v \in V$ is also associated with an attribute vector $\mathbf{x}_v \in \R^d$ and let $\mathbf{X} \in \R^{N \times d}$ represent the attribute vectors for all nodes in the graph. Let $\mathbf{A} \in \R^{N \times N}$ represent the graph adjacency matrix; here we assume that $\mathbf{A}$ is a symmetric and binary matrix such that $\mathbf{A}_{ij} \in \{0, 1\}$, where $\mathbf{A}_{ij}=1$ if there is an edge between nodes $i$ and $j$, i.e., $(v_i, v_j) \in E$, and $A_{ij}=0$ otherwise. Also, let $\mathbf{D}$ represent the diagonal degree matrix such that $\mathbf{D}_{ii} = \sum_{j=0}^{N-1}\mathbf{A}_{ij}$. 

Typical GNNs learn node representations via a neighborhood aggregation function. Assuming a GNN with $K$ layers, we define such a neighborhood aggregation function centred on node $v$ at layer $l$ as follows,
\begin{equation}
    \mathbf{h}^{(l)}_v = \activ{l} \left( f\left( g\left(\mathbf{h}^{(l-1)}_v, \mathbf{h}^{(l-1)}_u~\forall u \in \neigh{v} \right) \right) \right),
    \label{eq:gnn_layer}
\end{equation}
where $\neigh{v}$ is the set of node $v$'s neighbors in the graph, $g$ is an aggregation function, $f$ is a linear transformation that could be the identity function, and $\activ{l}$ is a non-linear function applied element-wise. Let $\mathbf{H}^{(l)} \in \R^{N \times d^{(l)}}$ the representations for all nodes at the $l$-th  layer with output dimension $d^{(l)}$; we set $\mathbf{H}^{(0)} \defeq \mathbf{X}$. and $d^{(0)} \defeq d$. A common aggregation function $g$ that calculates the weighted average of the node features where the weights are a deterministic function of the node degrees is $\hat{\mathbf{A}}\mathbf{H}$ as proposed by~\citet{kipf2016semi}. Here $\hat{\mathbf{A}}$ represents the twice normalized adjacency matrix with self loops given by $\hat{\mathbf{A}} = \hat{\mathbf{D}}^{-1/2}(\mathbf{A}+\mathbf{I})\hat{\mathbf{D}}^{-1/2}$ where $\hat{\mathbf{D}}$ is the degree matrix for $\mathbf{A}+\mathbf{I}$ and $\mathbf{I} \in \R^{N \times N}$ is the identity matrix. Substituting this aggregation function in \cref{eq:gnn_layer}, specifying $f$ to be a linear projection with weights $\mathbf{W}$ 
and defining the matrix $\mbOmega$ such that $\Omega_{ij} \defeq \hat{A}_{ij}$,
gives rise to the graph convolutional layer of \citet{kipf2016semi},
\begin{equation}
    \mathbf{H}^{(l)}=\activ{l}(\mbOmega \mathbf{H}^{(l-1)}\mathbf{W}^{(l)}), 
    \label{eq:gcn_layer}
\end{equation}
where, as before, $\activ{l}$ is  a non-linear function, typically the element-wise rectified linear unit (ReLU) activation \citep{nair2010rectified}. 
Many other aggregation functions have been proposed, most notably the sampled mean aggregator in GraphSAGE \citep{hamilton2017inductive} and the attention-based weighted mean aggregator in graph attention networks \citep[GAT][]{velickovic2018graph}. 
In our work, we employ GAT-based graph convolutional layers, as they have been shown by \citet{dwivedi2020benchmarking} to be more expressive than the graph convolutional network (GCN) architecture of \citet{kipf2016semi}. In this case we make 
$\Omega_{ij} \defeq \omega_{ij}$ with
\begin{equation}
\omega_{ij} = \frac{\exp \left(\leakyrelu\left(\mathbf{\mbalpha}^T[\mathbf{W}\mathbf{h}_i\|\mathbf{W}\mathbf{h}_j]\right)\right)}{\sum_{k \in \neigh{i}}\exp\left(\leakyrelu\left(\mathbf{\mbalpha}^T[\mathbf{W}\mathbf{h}_i\|\mathbf{W}\mathbf{h}_k]\right)\right)}
\label{eq:gat}
\end{equation}
for $j \in \neigh{i}$, where $\neigh{i}$, as before, is the set of node $i$'s neighbors; $\mathbf{\mbalpha}$ and $\mathbf{W}$ are trainable weight vector and matrix respectively and $\|$ is the concatenation operation.

\subsection{Node Property Prediction}
\Cref{eq:gcn_layer} 
is a realization of \cref{eq:gnn_layer} and constitutes the so-called spatial graph convolutional layer. More than one such layers can be stacked together to define GNNs.
When paired with a task-specific loss function, these GNNs can be used to learn node representations in a semi-supervised setting using full-batch gradient descent. For example, in semi-supervised node classification, it is customary to use the row-wise softmax function at the output layer along with the cross-entropy loss over the training (labeled) nodes.

\subsection{Graph Property Prediction}
In the graph property prediction setting, we are given a set of $M$ graphs $\bar{G}=\{G_0, G_1, ..., G_{M-1}\}$ and corresponding properties (labels) $\mbY = \{\mby_0, \mby_1, ..., \mby_{M-1}\}$. The goal is to learn a function 
that maps a graph to its properties. The standard approach is to first learn node representations using a $K$-layer GNN followed by a readout function that outputs a graph-level vector representation. This graph-level representation can be used as input to a classifier or regressor. The readout function for a graph $G$ is generally defined as,
\begin{equation}
    \mathbf{h}_{G}=r(\mathbf{h}^{(K-1)}_v \g v \in G),
    \label{eq:redout}
\end{equation}
where $\mathbf{h}_G \in \R^{\dgraph}$ such that $\dgraph$ is the dimensionality of the graph-level representation vectors. Note that Equation~\ref{eq:redout} aggregates representations from all nodes in the graph.
\begin{figure*}[t]
	\begin{center}
	\includegraphics[height=7.5cm]{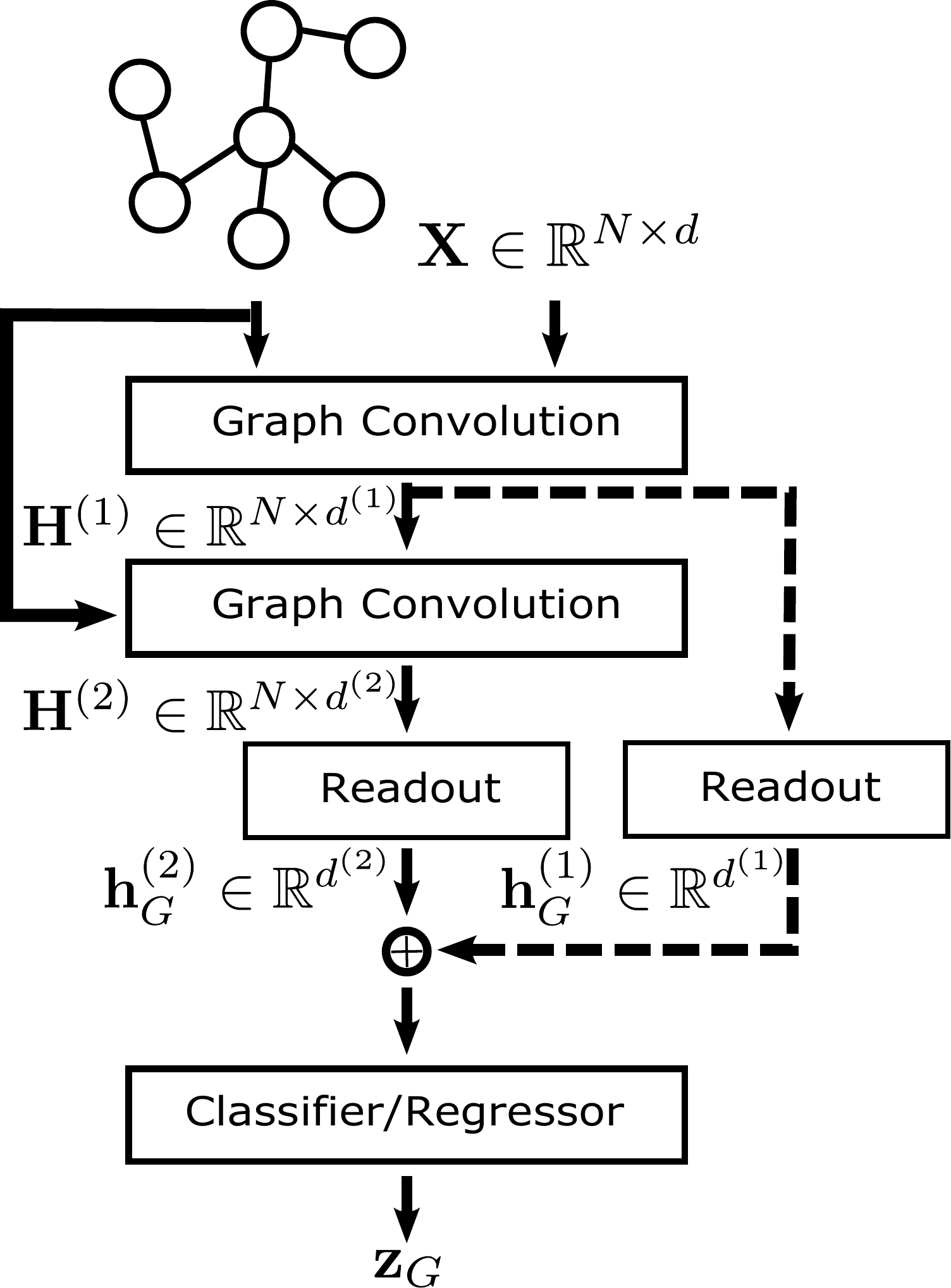}
    \hskip 0.3in
    \includegraphics[height=7.5cm]{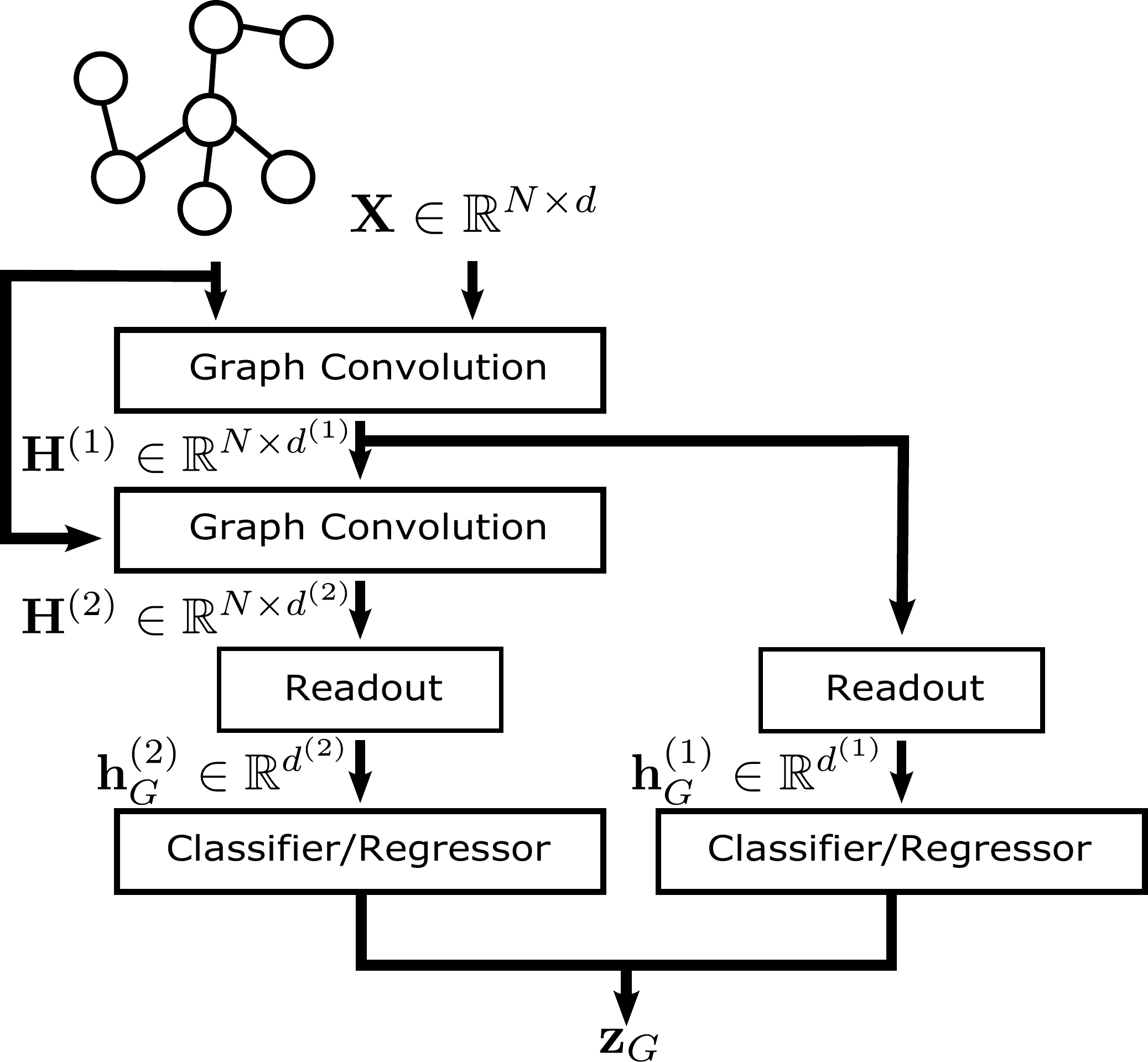}
    \caption{GNN architectures for graph property prediction. \textit{Left}: the standard architecture using a single readout layer after the last graph convolution but also shown with optional jump connections (dashed lines). \textit{Right}: the proposed architecture with deep supervision. For node property prediction the readout layers are removed from their corresponding architecture.}
    \end{center}
	\label{fig:gnn_architectures}
	\vskip -0.15in
\end{figure*}

Figure~\ref{fig:gnn_architectures} (left) shows a diagram of the standard GNN architecture with optional jump connections; for node property prediction tasks the architecture is the same but with the Readout layers removed. Jump connections can be applied at the node level, i.e., concatenate node representations output from each convolutional layer, or at the graph-level as shown in Figure~\ref{fig:gnn_architectures}. Furthermore, we include a multi-layer Perceptron (MLP) as the classifier/regressor so that the network can be trained end-to-end using stochastic gradient descent. The MLP is optional when using the standard architecture but necessary when employing jump connections. Given a suitable loss function such as the cross-entropy for classification or the root mean squared error (RMSE) for regression, we can train predictive models in a supervised setting for graph-level tasks and semi-supervised setting for node-level tasks.


%
\section{Deeply-supervised Graph Neural Networks}
\label{sec:ds_gnns}

Deeply-supervised nets \citep[DSNs,][]{pmlr-v38-lee15a} were proposed as a solution to several problems in training deep neural networks. 
By using companion objective functions attached to the output of each hidden layer, DSNs tackle the issue of vanishing gradients. Furthermore, in standard neural networks with shallow architectures, deep supervision operates as a regularizer of the loss at the last hidden layer. Lastly, and more importantly, for deep networks, it encourages the estimation of discriminative features at all network layers \citep{pmlr-v38-lee15a}. Therefore, inspired by this work, we introduce deeply supervised graph neural networks (DSGNNs), i.e., graph neural network architectures trained with deep supervision. Thus, we hypothesize that DSGNNs are resilient to over-smoothing and test this hypothesis by evaluating and analyzing their performance in training shallow and deep networks in \cref{sec:empirical_evaluation}.

Once we have defined node-level representations, as described in \cref{sec:graph_convolutional_networks}, our first step to construct and train DSGNNs is to compute graph-level representations at each layer using $\mbh_G^{(l)} = r(\mbH^{(l)})$, where $\mbH^{(l)}$ is obtained using the recurrent relation in \cref{eq:gcn_layer}.  As before, $\readout(\cdot)$ is a readout (or pooling) function. Simple examples of readout functions are the mean and the maximum of the features across all the nodes in the graph. This is followed by a linear layer and (potentially) a non-linearity that computes the output for each layer, $\mbz_G^{(l)} = h_G(\mbh_G^{(l)} \mbW_G^{(l)})$. For example, $h_G(\cdot)$ can be the softmax function or the identity function for classification or regression, respectively. Finally, given a loss function over our true and predicted outputs $\{(\mby_G, \mbz_G^{l})\}$ we learn all model parameters by optimizing the average loss function across all layers. 
\subsection{Graph Classification with a 2-Layer Network}
As an illustrative example, here we consider a graph classification problem with $C$ classes using a $2$-layer GAT model 
as shown in \cref{fig:gnn_architectures} (right). 
See Section \ref{sec:ds_for_node_classification} in the Appendix for an example of the node classification setting. 

\textbf{(i) Layer-dependent graph features}: 
We first compute, for each graph, layer-dependent graph features as: 
\begin{align}
   \mbH^{(1)} = \relu(\mbOmega \mbX \mbW^{(1)}), \quad   \mbh_G^{(1)} = \max (\mbH^{(1)}), \\
   \mbH^{(2)} = \relu(\mbOmega \mbH^{(1)} \mbW^{(2)}), 
   \quad \mbh_G^{(2)} = \max (\mbH^{(2)}),
\end{align}
where the $\relu$ activations are element-wise and the $\max$ readouts operate across rows.

\textbf{(ii) Layer-dependent outputs}: We then compute the outputs for each layer as:
\begin{equation}
    \mbz_G^{(l)} = \softmax(\mbh_G^{(l)} \mbW_G^{(l)}), 
    \ \quad l = 1, 2,
\end{equation}
where we note the new parameters $\{\mbW_G^{(l)}\}$, which are different from the previous weight matrices $\{\mbW^{(l)}\}$.  
%

\textbf{(iii) Layer-dependent losses}: We now compute the cross-entropy loss for each layer:
\begin{equation}
    \cL^{(l)}_{\bar{G}} = -\sum_{g \in G_L} \sum_{c=0}^{C-1} \mathbf{Y}_{g, c} \log(\mbZ^{(l)}_{g, c}), \quad l=1,2,
    \label{eq:cross_entropy}
\end{equation}
where $G_L \subseteq \bar{G}$ is the set of training graphs, $\mbZ^{(l)}_{g,c}$ is the predicted probability for class $c$ and graph $g$, and $\mathbf{Y}_{g,c}$ is the corresponding ground truth label.

\textbf{(iv) Total loss}:  
The DSGNN loss is the mean of the losses of all predictive layers, for $K=2$, we have:
\begin{equation}
    \cL_{\bar{G}} = 
    \frac{1}{K}\sum_{k=1}^{K} \cL^{(k)}_{\bar{G}},
    \label{eq:loss_ds}
\end{equation}
where each of the individual losses are given by \cref{eq:cross_entropy}.
We estimate the model parameters using gradient-based optimization so as to minimize the total loss in \cref{eq:loss_ds}. 
Unlike \citet{pmlr-v38-lee15a}, we do not decay the contribution of the surrogate losses as a function of the training epoch. Consequently,  at prediction time we average the outputs from all classifiers and then apply the softmax function to make a single prediction for each input graph. 

\subsection{Advantages of Deep Supervision} 
As mentioned before, over-smoothing leads to node representations with low discriminative power at the last GNN layer. This hinders the deep GNN's ability to perform well on predictive tasks. DSGNNs circumvent this issue as the learned node representations from all hidden layers inform the final decision. The distributed loss encourages node representations learned at all hidden layers to be discriminative such that network predictions do not rely only on the discriminative power of the last layer's representations.

Furthermore, deep supervision increases the number of model parameters linearly to the number of MLP layers. Consider a classification model with $K$ hidden layers, $\dgraph$ dimensional graph-level representations, and a single layer MLP. If the number of classes is $C$, then a DSGNN model requires $(K-1) \times \dgraph \times C$ parameters more than a standard GNN. 

\section{Empirical Evaluation}
\label{sec:empirical_evaluation}

We aim to empirically evaluate the performance of DSGNNs on a number of challenging graph and node classification and regression tasks. 
We investigate empirically if the addition of deep supervision provides an advantage over the standard GNN and JKNet \citep{jumping-knowledge-networks-xu18c} architectures shown in  \cref{fig:gnn_architectures}. 

We implemented\footnote{We will release the source code upon publication acceptance} the standard GNN, JKNet, and DSGNN architectures using PyTorch and the Deep Graph Library  \citep[DGL,][]{wang2019dgl}. The version of the datasets we use is that available via DGL\footnote{\url{https://github.com/dmlc/dgl}} and DGL-LifeSci\footnote{\url{https://github.com/awslabs/dgl-lifesci}}. All experiments were run on a workstation with $8$GB of RAM, Nvidia Telsa P100 GPU, and Intel Xeon processor.
\begin{table}[t]
    \caption{Graph regression and classification performance for the standard GNN, JKNet, and DSGNN architectures. The performance metric is mean test RMSE for \esol and \lipo and mean test accuracy for \enzymes calculated using $10$ repeats of $10$-fold cross validation. Standard deviation is given in parenthesis and the model depth that achieved the best performance in square brackets. Bold and underline indicate the best and second best models for each dataset.}
    \vskip 0.15in
    \begin{center}
    \begin{tabular}{llll}
        \toprule
        Model &  \esol & \lipo & \enzymes \\ 
              &   \multicolumn{2}{c}{RMSE $\downarrow$} & \multicolumn{1}{c}{Accuracy $\uparrow$} \\
        \midrule
        GNN     &  \underline{0.726 (0.063) [6]} & \underline{0.618(0.033) [8]} & \underline{64.1(6.8) [2]} \\
        JKNet   &  0.728 (0.074) [8]             & 0.633 (0.035) [10]           & \textbf{65.7 (5.8) [2]}  \\ 
        DSGNN   &  \textbf{0.694 (0.065) [16]}   & \textbf{0.594 (0.033) [16]}  & 63.3 (7.7) [2] \\ 
        \bottomrule
    \end{tabular}
    \label{tab:results_summary_graph_datasets}
    \end{center}
    \vskip -0.15in
\end{table}
\begin{table}[t]
    \caption{Node classification performance for  the standard GNN, JKNet, and DSGNN architectures with and without PairNorm (PN). The performance metric is mean test accuracy calculated  over $20$ repeats of fixed train/val/test splits. Standard deviation is given in parenthesis and the model depth that achieved the best performance in square brackets. Bold and underline indicate the best and second best performing models for each dataset.}
    \vskip 0.15in
    \begin{center}
    \begin{tabular}{l  l  l  l}
        \toprule
        Model &  \cora & \citeseer & \pubmed \\ 
              &  \multicolumn{3}{c}{Accuracy $\uparrow$} \\
        \midrule
        GNN        &   \textbf{82.6 (0.6) [2]}        &  \textbf{71.1 (00.6) [2]}    & \underline{77.2 (0.5) [9]} \\
        JKNet      &   \underline{81.4 (0.6) [3]}     &  68.5 (0.4) [2]              & 76.9 (0.9) [11]  \\ 
        DSGNN      &   81.1 (1.0) [4]                 &  \underline{69.9 (0.4) [3]}  & \textbf{77.5 (0.5) [12]} \\ 
        GNN-PN     &   77.9 (0.4) [2]                 &  68.0 (0.7) [3]              & 75.5 (00.7) [15] \\
        DSGNN-PN   &   73.1 (0.8) [7]                 &  59.4 (1.6) [2]              & 75.9 (00.5) [7]  \\ 
        \bottomrule
    \end{tabular}
    \label{tab:results_summary_node_datasets}
    \end{center}
    \vskip -0.15in
\end{table}
\begin{table}[t]
    \caption{\textbf{Missing features setting} node classification performance comparison between the standard GNN, JKNet and DSGNN architectures with and without PairNorm (PN). Results shown are mean test accuracy and standard deviation over $20$ repeats of fixed train/val/test splits. In parenthesis we indicate the model depth that achieved the highest accuracy. Bold and underline indicate the best and second best performing models.}
    \vskip 0.15in
    \begin{center}
    \begin{tabular}{l  l  l  l}
        \toprule
        Model &  \cora & \citeseer & \pubmed \\ 
              &  \multicolumn{3}{c}{Accuracy $\uparrow$} \\
        \midrule
        GNN        &  \textbf{77.5 (0.8) [10]}     &   \textbf{62.8 (0.7) [2]}       &  \underline{76.8 (0.7) [9]} \\
        JKNet      &  74.9 (1.0) [15]              &   61.8 (0.8) [2]                &  76.4 (0.7) [9]  \\ 
        DSGNN      &  \underline{76.8 (0.8) [11]}  &   61.0 (1.0) [2]                &  \textbf{0.771 (0.4) [10]} \\ 
        GNN-PN     &  75.8 (0.4) [6]               &   \underline{62.1 (0.6) [4]}    &  75.0 (0.7) [15] \\
        DSGNN-PN   &  73.5 (0.9) [15]              &   52.8 (1.2) [9]                &  74.7 (0.9) [25] \\ 
        \bottomrule
    \end{tabular}
    \label{tab:results_summary_node_datasets_mv}
    \end{center}
    \vskip -0.15in
\end{table}

\subsection{Datasets and Experimental Set-up}
\label{sec:datasets}

DSGNN is a general architecture such that it can use any combination of graph convolutional and readout layers. We focus the empirical evaluation on a small number of representative methods. For graph convolutions we use graph attention networks \citep[GAT,][]{velickovic2018graph} with multi-head attention
We average or concatenate the outputs of the attention heads (we treat this operation as a hyper-parameter) and use the resulting node vectors as input to the next layer.

For DSGNN and JKNet, the last GAT layer is followed by a fully connected layer with an activation suitable for the downstream task, e.g., softmax for classification. The linear layer is necessary to map the GAT layer representations to the correct dimension for the downstream task. So, a DSGNN or JKNet model with $K$ layers comprises of $K-1$ GAT layers and one linear layer. For a standard GNN model, the last layer is also GAT following \citet{velickovic2018graph} such that a $K$-layer model comprises of $K$ GAT layers.

We used the evaluation protocol proposed by~\citet{errica2019fair} and present results for six benchmark datasets. Of these, three are graph tasks and three are node tasks. We include detailed dataset statistics in Table \ref{tab:datasets} in the Appendix.

\subsubsection{Graph Property Prediction}
\label{sec:graph_regression_datasets}

The graph property prediction datasets are from biochemistry where graphs represent molecules. The task is to predict molecular properties. We base our empirical evaluation on three datasets which are ESOL from \citet{delaney2004esol}, Lipophilicity from \citet{gaulton2017chembl} and \enzymes from \citet{schomburg2004brenda}. \enzymes is a graph classification task whereas \esol and \lipo are regression tasks. We use $10$-fold cross validation and repeat each experiment $10$ times. We optimize the root mean square error (RMSE) for regression and the cross-entropy loss for classification.

For all architectures, we set all hidden layers to size $512$ with $4$ attention heads (total $2048$ features). All layers are followed with a $\relu$ activation~\citep{nair2010rectified} except the last one; the last layer either uses $\softmax$ activation or no activation for classification and regression tasks respectively. We varied model depth in the range $\{ 2,4,6,\ldots,20 \}$. For readout, we use the non-parametric max function. We perform grid search for learning rate: \{$0.01$, $0.001$, $0.0001$\} and weight decay: \{$0.001$, $0.0001$\}. We use batch size $64$ and train for a maximum $500$ epochs using mini-batch SGD with momentum set to $0.9$.
\subsubsection{Node Property Prediction}
\label{sec:node_classification_datasets}

The node classification datasets are the citation networks \cora, \citeseer and \pubmed from \citet{sen2008collective}. The task for all datasets is semi-supervised node classification in a regime with few labeled nodes. We used the splits from \citet{yang2016revisiting} and performance on the validation set for hyper-parameter selection. We repeat each experiment $20$ times. We optimize the cross-entropy loss and use accuracy for model selection and performance comparison.

We set all hidden layers to size $8$ with $8$ attention heads (total $64$ features) and $\elu$~\citep{clevert2015fast} activation for all GAT layers. We varied model depth in the range $\{ 2,3,\ldots,12,15,20,25 \}$. We performed grid search for the initial learning rate: \{$0.01, 0.002, 0.005$\}, weight decay: \{$0.0, 0.005, 0.0005$\}, and dropout (both feature and attention): \{$0.2, 0.5$\}. We trained all models using the Adam optimiser~\citep{kingma2014adam} for a maximum $1000$ epochs and decayed the learning rate by a factor of $0.5$ every $250$ epochs.
\begin{figure*}[t]
	\centering
	    \begin{tabular}{c}
        	\includegraphics[width=16.5cm]{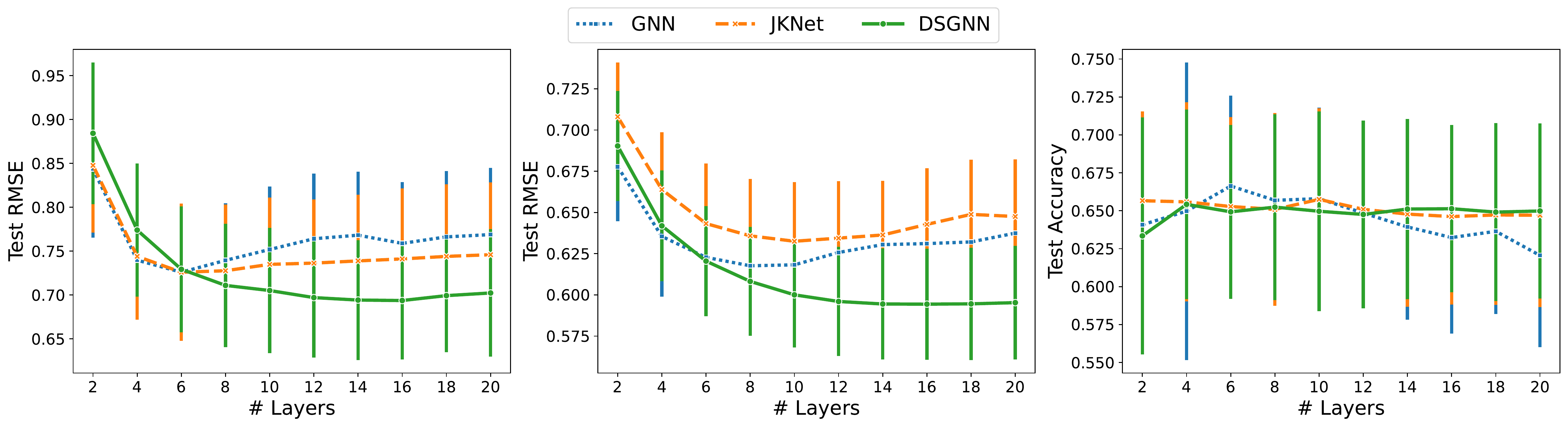} \\
        \end{tabular}
    \caption{Graph classification and regression performance of standard GNN, JKNet, and DSGNN architectures as a function of model depth. Results shown are for the \esol (left), \lipo (middle) and \enzymes (right) datasets. The performance metric for \esol and \lipo is test RMSE and for \enzymes test accuracy. All metrics are mean over $10$ runs with $1$ standard deviation error bars.}
	\label{fig:deep_gnn_graph_performance}
	\vskip -0.1in
\end{figure*}
\begin{figure*}[t]
	\centering
	    \begin{tabular}{c}
        	\includegraphics[width=16.5cm]{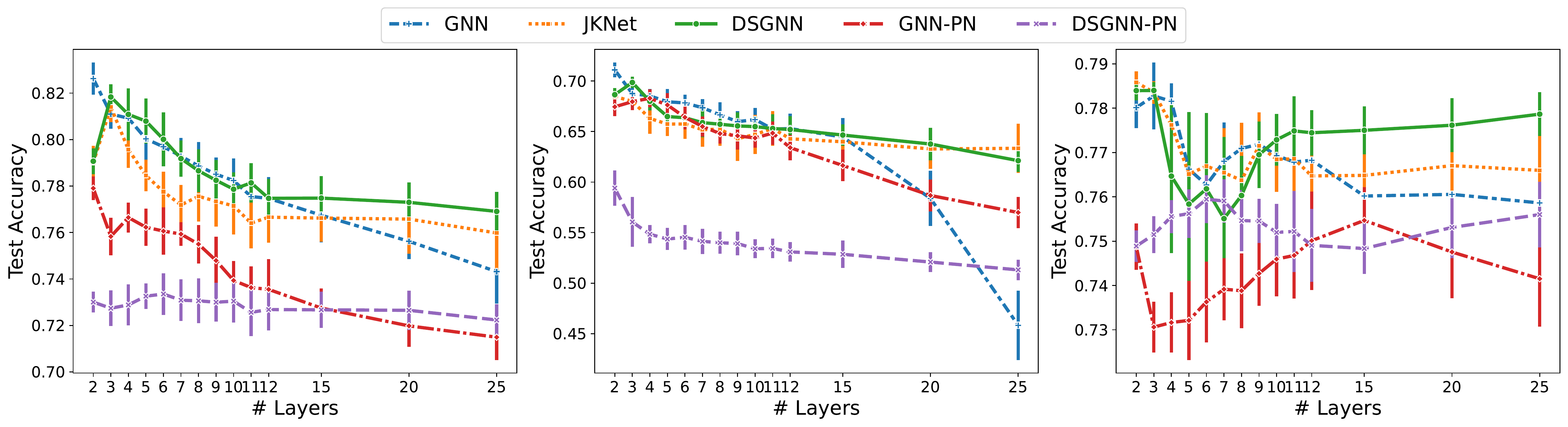} \\
        \end{tabular}
    \caption{Node classification performance comparison of standard GNN, JKNet, and DSGNN with and without PairNorm (PN) architectures as a function model depth. Results shown are for Cora (left), Citeseer (middle), and Pubmed (right) datasets. The performance metric is mean test over $20$ runs with $1$ standard deviation error bars.}
	\label{fig:deep_gnn_node_performance}
\end{figure*}
\begin{figure*}[t]
	\centering
	    \begin{tabular}{c}
        	\includegraphics[width=16.5cm]{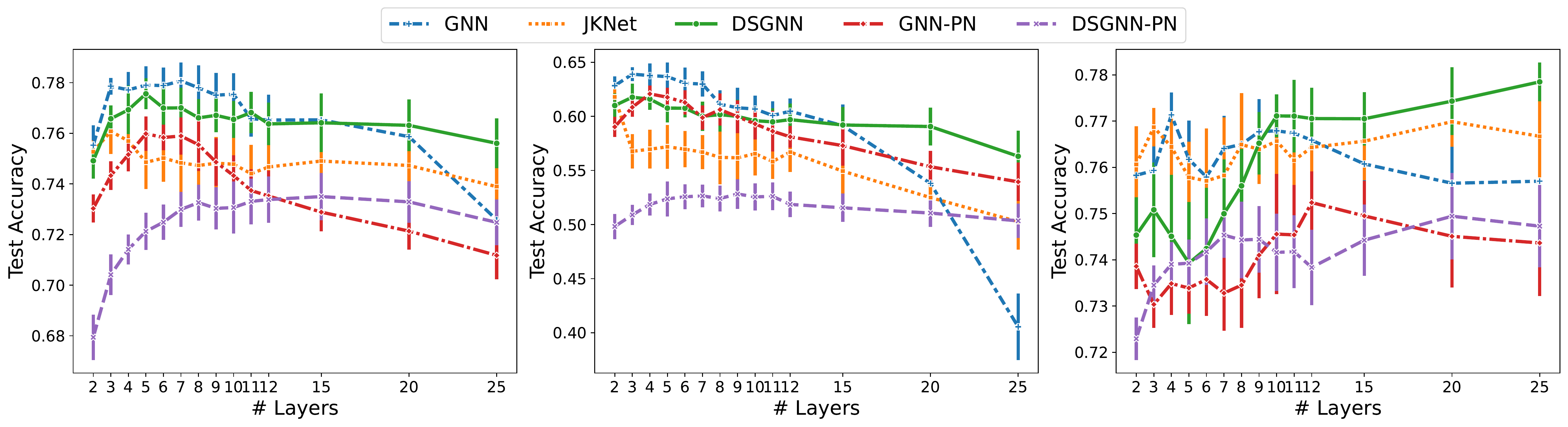} \\
        \end{tabular}
    \caption{\textbf{Missing features setting} node classification performance comparison of standard GNN, JKNet, and DSGNN with and without PairNorm (PN) architectures as a function model depth. Results shown are for Cora (left), Citeseer (middle), and Pubmed (right) datasets. The performance metric is mean test over $20$ runs with $1$ standard deviation error bars.}
	\label{fig:deep_gnn_node_mv_performance}
	\vskip -0.1in
\end{figure*}

\subsection{Results}
\label{sec:results}

Tables \ref{tab:results_summary_graph_datasets} and \ref{tab:results_summary_node_datasets} show the performance for each architecture and for the all datasets. Also shown for each model is the number of layers required to achieve the best performance. For each dataset, model selection across model depth is based on validation set performance and the tables report performance on the test set for the selected models.   

\subsubsection{Graph Regression and Classification}
\label{sec:results_graph_regression_and_classification}

We see in \cref{tab:results_summary_graph_datasets} that for \esol and \lipo, models enhanced with deep supervision achieved the best outcome. In addition, DSGNN performs best at a much larger depth than the other architectures. On the graph classification dataset (\enzymes), all models perform similarly with JKNet having a small advantage. For graph classification, all architectures performed best using only $2$ convolutional layers. 

Figure~\ref{fig:deep_gnn_graph_performance} shows the performance of each architecture as a function of model depth. For the regression datasets, all architectures benefit from increasing depth up to a point. For GNN models with more than $6$ and $8$ layers for \esol (left) and \lipo (middle) datasets respectively performance starts decreasing. Similarly for JKNet for models with more than $8$ and $10$ layers for \esol and \lipo respectively. On the other hand, with  the exception of the \enzymes dataset, DSGNN performance continues to improve for up to $16$ layers for both datasets before it flattens out. Our evidence suggests that for graph-level tasks 
DSGNNs can exploit larger model depth to achieve improved performance when compared to standard GNNs and JKNets.

\subsubsection{Node Classification}
\label{sec:results_node_classification}
Table~\ref{tab:results_summary_node_datasets} shows that the standard GNN architecture outperformed the others in two of the three node classification datasets, namely \cora and \citeseer. DSGNN demonstrated a small advantage on the larger \pubmed dataset. As previously reported in the literature, shallow architectures perform best on these citation networks. We observe the same since for the smaller \cora and \citeseer all architectures performed best with $2$ to $4$ layers. Only on \pubmed all architectures benefited from increased model depth. We attribute this to the graph's larger size where more graph convolutional layers allow for information from larger neighborhoods to inform the inferred node representations. 

It can be seen in Figure~\ref{fig:deep_gnn_node_performance} that for the smaller \cora and \citeseer, for all models performance degrades as model depth increases. As we noted earlier, this performance degradation has been attributed to the over-smoothing problem in GNNs. In our experiments, DSGNN demonstrated consistently higher resilience to over-smoothing than competing methods. 
\Cref{tab:results_summary_node_datasets_25} in the Appendix shows the performance of all architectures with $25$ layers and clearly indicates that DSGNN outperforms the standard GNN and JKNet for \cora and \pubmed. JKNet is best for \citeseer with DSGNN a close second. As expected DSGNN and JKNet are more resilient to over-smoothing as compared to the standard GNN with more than $12$ layers.  For all datasets and especially \citeseer, performance for the standard GNN degrades substantially as a function of model depth.

Lastly, we note that for all datasets the addition of pair normalization \citep[PairNorm,][]{Zhao2020PairNorm} to the standard GNN hurts performance. This finding is consistent with the results in \citet{Zhao2020PairNorm}. However, as we will see in \cref{sec:missing_features}, PairNorm can be beneficial in the missing feature setting. \cref{tab:results_summary_graph_datasets} shows that the performance of GNN with PN drops by approximately $4.7\%, 3.1\%$ on \cora and \citeseer respectively. DSGNN with PairNorm is the worst performing architecture across all node classification datasets. Consequently, we do not recommend the combination of deep supervision and PairNorm. 

\subsubsection{Node Classification with Missing Features}
\label{sec:missing_features}

\Citet{Zhao2020PairNorm} introduced the missing features setting for the node classification task and demonstrated that GNNs with PairNorm achieve the best results and for deeper models. In the missing features setting, a proportion of nodes in the validation and test sets have their feature vectors zeroed. This setting simulates the missing data scenario common in real-world applications. The missing data proportion can vary from $0\%$ where all nodes have known attributes and $100\%$ where all nodes in the validation and test sets have missing attributes. Here we consider the performance of standard GNN, JKNet, and DSGNN for the latter setting only.

Table~\ref{tab:results_summary_node_datasets_mv} and Figure~\ref{fig:deep_gnn_node_mv_performance} show the performance of the three architectures in the missing features setting. We note that in comparison to the results in Table~\ref{tab:results_summary_node_datasets} and excluding \citeseer, all models achieved their best performance at larger depth. Interestingly and in contrast to \citet{Zhao2020PairNorm}, we found that the standard GNN architecture performed best for the smaller \cora and \citeseer graphs. We attribute the standard GNN's good performance to our use of a model with high capacity ($8$ attention heads and $8$-dimensional embeddings for each head) as well as careful tuning of relevant hyper-parameters. \Citet{Zhao2020PairNorm} use a simpler model, e.g., one attention head, and do not tune important hyper-parameters such as learning rate, dropout and weight decay.

However, on the larger \pubmed dataset, DSGNN with $10$ layers achieves the highest test accuracy. A DSGNN model with $25$ layers as shown in Figure~\ref{fig:deep_gnn_node_mv_performance} and Table~\ref{tab:results_summary_node_datasets_25_mv} in the Appendix achieves the highest test accuracy even when compared to the $10$-layer DSGNN model; the latter was selected for inclusion in Table~\ref{tab:results_summary_node_datasets_mv} because it achieved the highest validation accuracy that we used for model selection across model depth. We provide additional analysis of DSGNN's ability to learn more discriminative node representations and alleviate over-smoothing in the Appendix Section \ref{sec:learned-representations}. We conclude that DSGNN is the more robust architecture to the over-smoothing problem in the missing feature setting and especially for larger graphs.

\section{Conclusion}
\label{sec:conclusion}

We introduced deeply-supervised graph neural networks (DSGNNs) and demonstrated their effectiveness in training high performing models for graph and node property prediction problems. DSGNNs are GNNs enhanced with deep supervision that introduce companion losses attached to the hidden layers guiding the learning algorithm to learn discriminative features at all model depths. 

DSGNNs overcome the over-smoothing problem in deep models achieving competitive performance when compared with standard GNNs enhanced with PairNorm and jump connections. We provided empirical evidence supporting this for both graph and node property prediction and in the missing feature setting. We found that combining deep supervision with PairNorm degrades model performance. DSGNNs are more resilient to the over-smoothing problem achieving substantially higher accuracy for deep models. In future work, we plan to investigate the application of DSGNNs on larger graphs where we expect deep supervision will be beneficial.

\bibliography{bib/references}

\begin{thebibliography}{37}
\providecommand{\natexlab}[1]{#1}
\providecommand{\url}[1]{\texttt{#1}}
\expandafter\ifx\csname urlstyle\endcsname\relax
  \providecommand{\doi}[1]{doi: #1}\else
  \providecommand{\doi}{doi: \begingroup \urlstyle{rm}\Url}\fi

\bibitem[Chen et~al.(2020)Chen, Lin, Li, Li, Zhou, and Sun]{chen2020measuring}
Deli Chen, Yankai Lin, Wei Li, Peng Li, Jie Zhou, and Xu~Sun.
\newblock Measuring and relieving the over-smoothing problem for graph neural
  networks from the topological view.
\newblock In \emph{Proceedings of the AAAI Conference on Artificial
  Intelligence}, volume~34, pages 3438--3445, 2020.

\bibitem[Chiang et~al.(2019)Chiang, Liu, Si, Li, Bengio, and
  Hsieh]{chiang2019cluster}
Wei-Lin Chiang, Xuanqing Liu, Si~Si, Yang Li, Samy Bengio, and Cho-Jui Hsieh.
\newblock Cluster-gcn: An efficient algorithm for training deep and large graph
  convolutional networks.
\newblock In \emph{Proceedings of the 25th ACM SIGKDD International Conference
  on Knowledge Discovery \& Data Mining}, pages 257--266, 2019.

\bibitem[Clevert et~al.(2015)Clevert, Unterthiner, and
  Hochreiter]{clevert2015fast}
Djork-Arn{\'e} Clevert, Thomas Unterthiner, and Sepp Hochreiter.
\newblock Fast and accurate deep network learning by exponential linear units
  (elus).
\newblock \emph{arXiv preprint arXiv:1511.07289}, 2015.

\bibitem[Delaney(2004)]{delaney2004esol}
John~S Delaney.
\newblock {ESOL}: estimating aqueous solubility directly from molecular
  structure.
\newblock \emph{Journal of chemical information and computer sciences},
  44\penalty0 (3):\penalty0 1000--1005, 2004.

\bibitem[Dwivedi et~al.(2020)Dwivedi, Joshi, Laurent, Bengio, and
  Bresson]{dwivedi2020benchmarking}
Vijay~Prakash Dwivedi, Chaitanya~K Joshi, Thomas Laurent, Yoshua Bengio, and
  Xavier Bresson.
\newblock Benchmarking graph neural networks.
\newblock \emph{arXiv preprint arXiv:2003.00982}, 2020.

\bibitem[Errica et~al.(2019)Errica, Podda, Bacciu, and Micheli]{errica2019fair}
Federico Errica, Marco Podda, Davide Bacciu, and Alessio Micheli.
\newblock A fair comparison of graph neural networks for graph classification.
\newblock \emph{arXiv preprint arXiv:1912.09893}, 2019.

\bibitem[Gao and Ji(2019)]{graph-u-nets-gao19a}
Hongyang Gao and Shuiwang Ji.
\newblock Graph u-nets.
\newblock In Kamalika Chaudhuri and Ruslan Salakhutdinov, editors,
  \emph{Proceedings of the 36th International Conference on Machine Learning},
  volume~97 of \emph{Proceedings of Machine Learning Research}, pages
  2083--2092. PMLR, 09--15 Jun 2019.
\newblock URL \url{http://proceedings.mlr.press/v97/gao19a.html}.

\bibitem[Gaudelet et~al.(2021)Gaudelet, Day, Jamasb, Soman, Regep, Liu, Hayter,
  Vickers, Roberts, Tang, Roblin, Blundell, Bronstein, and
  Taylor-King]{10.1093/bib/bbab159}
Thomas Gaudelet, Ben Day, Arian~R Jamasb, Jyothish Soman, Cristian Regep,
  Gertrude Liu, Jeremy B~R Hayter, Richard Vickers, Charles Roberts, Jian Tang,
  David Roblin, Tom~L Blundell, Michael~M Bronstein, and Jake~P Taylor-King.
\newblock {Utilizing graph machine learning within drug discovery and
  development}.
\newblock \emph{Briefings in Bioinformatics}, 22\penalty0 (6), 05 2021.
\newblock ISSN 1477-4054.
\newblock \doi{10.1093/bib/bbab159}.
\newblock URL \url{https://doi.org/10.1093/bib/bbab159}.
\newblock bbab159.

\bibitem[Gaulton et~al.(2017)Gaulton, Hersey, Nowotka, Bento, Chambers, Mendez,
  Mutowo, Atkinson, Bellis, Cibri{\'a}n-Uhalte, et~al.]{gaulton2017chembl}
Anna Gaulton, Anne Hersey, Micha{\l} Nowotka, A~Patricia Bento, Jon Chambers,
  David Mendez, Prudence Mutowo, Francis Atkinson, Louisa~J Bellis, Elena
  Cibri{\'a}n-Uhalte, et~al.
\newblock The {ChEMBL} database in 2017.
\newblock \emph{Nucleic acids research}, 45\penalty0 (D1):\penalty0 D945--D954,
  2017.

\bibitem[Hamilton et~al.(2017)Hamilton, Ying, and
  Leskovec]{hamilton2017inductive}
Will Hamilton, Zhitao Ying, and Jure Leskovec.
\newblock Inductive representation learning on large graphs.
\newblock In \emph{Advances in Neural Information Processing Systems}, pages
  1024--1034, 2017.

\bibitem[Hamilton(2020)]{hamilton-grl-book-2020}
William~L. Hamilton.
\newblock Graph representation learning.
\newblock \emph{Synthesis Lectures on Artificial Intelligence and Machine
  Learning}, 14\penalty0 (3):\penalty0 1--159, 2020.

\bibitem[Kingma and Ba(2014)]{kingma2014adam}
Diederik~P Kingma and Jimmy Ba.
\newblock Adam: A method for stochastic optimization.
\newblock \emph{arXiv preprint arXiv:1412.6980}, 2014.

\bibitem[Kipf and Welling(2017)]{kipf2016semi}
Thomas~N Kipf and Max Welling.
\newblock {S}emi-{S}upervised {C}lassification with {G}raph {C}onvolutional
  {N}etworks.
\newblock In \emph{International Conference on Learning Representations}, 2017.

\bibitem[Klicpera et~al.(2018)Klicpera, Bojchevski, and
  G{\"u}nnemann]{klicpera2018predict}
Johannes Klicpera, Aleksandar Bojchevski, and Stephan G{\"u}nnemann.
\newblock Predict then propagate: Combining neural networks with personalized
  pagerank for classification on graphs.
\newblock In \emph{International Conference on Learning Representations}, 2018.

\bibitem[Klicpera et~al.(2019)Klicpera, Bojchevski, and
  Günnemann]{klicpera2018combining}
Johannes Klicpera, Aleksandar Bojchevski, and Stephan Günnemann.
\newblock Combining neural networks with personalized pagerank for
  classification on graphs.
\newblock In \emph{International Conference on Learning Representations}, 2019.
\newblock URL \url{https://openreview.net/forum?id=H1gL-2A9Ym}.

\bibitem[Lee et~al.(2015)Lee, Xie, Gallagher, Zhang, and Tu]{pmlr-v38-lee15a}
Chen-Yu Lee, Saining Xie, Patrick Gallagher, Zhengyou Zhang, and Zhuowen Tu.
\newblock {Deeply-Supervised Nets}.
\newblock In Guy Lebanon and S.~V.~N. Vishwanathan, editors, \emph{Proceedings
  of the Eighteenth International Conference on Artificial Intelligence and
  Statistics}, volume~38 of \emph{Proceedings of Machine Learning Research},
  pages 562--570, San Diego, California, USA, 09--12 May 2015. PMLR.
\newblock URL \url{http://proceedings.mlr.press/v38/lee15a.html}.

\bibitem[Lee et~al.(2019)Lee, Lee, and Kang]{sag_pool}
Junhyun Lee, Inyeop Lee, and Jaewoo Kang.
\newblock Self-attention graph pooling.
\newblock In Kamalika Chaudhuri and Ruslan Salakhutdinov, editors,
  \emph{Proceedings of the 36th International Conference on Machine Learning},
  volume~97 of \emph{Proceedings of Machine Learning Research}, pages
  3734--3743. PMLR, 09--15 Jun 2019.
\newblock URL \url{http://proceedings.mlr.press/v97/lee19c.html}.

\bibitem[Li et~al.(2018)Li, Han, and Wu]{Li_Han_Wu_2018}
Qimai Li, Zhichao Han, and Xiao-ming Wu.
\newblock Deeper insights into graph convolutional networks for semi-supervised
  learning.
\newblock \emph{Proceedings of the AAAI Conference on Artificial Intelligence},
  32\penalty0 (1), Apr. 2018.
\newblock URL \url{https://ojs.aaai.org/index.php/AAAI/article/view/11604}.

\bibitem[Liu et~al.(2020)Liu, Gao, and Ji]{deep_gnn_meng20}
Meng Liu, Hongyang Gao, and Shuiwang Ji.
\newblock Towards deeper graph neural networks.
\newblock In \emph{Proceedings of the 26th ACM SIGKDD International Conference
  on Knowledge Discovery and Data Mining}, page 338–348, New York, NY, USA,
  2020. Association for Computing Machinery.
\newblock ISBN 9781450379984.
\newblock \doi{10.1145/3394486.3403076}.
\newblock URL \url{https://doi.org/10.1145/3394486.3403076}.

\bibitem[Nair and Hinton(2010)]{nair2010rectified}
Vinod Nair and Geoffrey~E Hinton.
\newblock Rectified linear units improve restricted boltzmann machines.
\newblock In \emph{ICML}, 2010.

\bibitem[Scarselli et~al.(2009)Scarselli, Gori, Tsoi, Hagenbuchner, and
  Monfardini]{scarselli_gnn}
Franco Scarselli, Marco Gori, Ah~Chung Tsoi, Markus Hagenbuchner, and Gabriele
  Monfardini.
\newblock The graph neural network model.
\newblock \emph{IEEE Transactions on Neural Networks}, 20\penalty0
  (1):\penalty0 61--80, 2009.
\newblock \doi{10.1109/TNN.2008.2005605}.

\bibitem[Schomburg et~al.(2004)Schomburg, Chang, Ebeling, Gremse, Heldt, Huhn,
  and Schomburg]{schomburg2004brenda}
Ida Schomburg, Antje Chang, Christian Ebeling, Marion Gremse, Christian Heldt,
  Gregor Huhn, and Dietmar Schomburg.
\newblock {BRENDA}, the enzyme database: updates and major new developments.
\newblock \emph{Nucleic acids research}, 32\penalty0 (suppl\_1):\penalty0
  D431--D433, 2004.

\bibitem[Sen et~al.(2008)Sen, Namata, Bilgic, Getoor, Galligher, and
  Eliassi-Rad]{sen2008collective}
Prithviraj Sen, Galileo Namata, Mustafa Bilgic, Lise Getoor, Brian Galligher,
  and Tina Eliassi-Rad.
\newblock Collective classification in network data.
\newblock \emph{AI magazine}, 29\penalty0 (3):\penalty0 93--93, 2008.

\bibitem[van~der Maaten and Hinton(2008)]{JMLR:v9:vandermaaten08a}
Laurens van~der Maaten and Geoffrey Hinton.
\newblock Visualizing data using t-sne.
\newblock \emph{Journal of Machine Learning Research}, 9\penalty0
  (86):\penalty0 2579--2605, 2008.
\newblock URL \url{http://jmlr.org/papers/v9/vandermaaten08a.html}.

\bibitem[Veli{\v{c}}kovi{\'{c}} et~al.(2018)Veli{\v{c}}kovi{\'{c}}, Cucurull,
  Casanova, Romero, Li{\`{o}}, and Bengio]{velickovic2018graph}
Petar Veli{\v{c}}kovi{\'{c}}, Guillem Cucurull, Arantxa Casanova, Adriana
  Romero, Pietro Li{\`{o}}, and Yoshua Bengio.
\newblock {Graph Attention Networks}.
\newblock In \emph{International Conference on Learning Representations}, 2018.

\bibitem[Wang et~al.(2019)Wang, Zheng, Ye, Gan, Li, Song, Zhou, Ma, Yu, Gai,
  Xiao, He, Karypis, Li, and Zhang]{wang2019dgl}
Minjie Wang, Da~Zheng, Zihao Ye, Quan Gan, Mufei Li, Xiang Song, Jinjing Zhou,
  Chao Ma, Lingfan Yu, Yu~Gai, Tianjun Xiao, Tong He, George Karypis, Jinyang
  Li, and Zheng Zhang.
\newblock Deep graph library: A graph-centric, highly-performant package for
  graph neural networks.
\newblock \emph{arXiv preprint arXiv:1909.01315}, 2019.

\bibitem[Wu et~al.(2019)Wu, Souza, Zhang, Fifty, Yu, and
  Weinberger]{wu2019simplifying}
Felix Wu, Amauri Souza, Tianyi Zhang, Christopher Fifty, Tao Yu, and Kilian
  Weinberger.
\newblock Simplifying graph convolutional networks.
\newblock In \emph{International Conference on Machine Learning}, pages
  6861--6871, 2019.

\bibitem[Wu and He(2018)]{wu2018group}
Yuxin Wu and Kaiming He.
\newblock Group normalization.
\newblock In \emph{Proceedings of the European conference on computer vision
  (ECCV)}, pages 3--19, 2018.

\bibitem[Xu et~al.(2018{\natexlab{a}})Xu, Hu, Leskovec, and
  Jegelka]{xu2018powerful}
Keyulu Xu, Weihua Hu, Jure Leskovec, and Stefanie Jegelka.
\newblock How powerful are graph neural networks?
\newblock In \emph{International Conference on Learning Representations},
  2018{\natexlab{a}}.

\bibitem[Xu et~al.(2018{\natexlab{b}})Xu, Li, Tian, Sonobe, Kawarabayashi, and
  Jegelka]{jumping-knowledge-networks-xu18c}
Keyulu Xu, Chengtao Li, Yonglong Tian, Tomohiro Sonobe, Ken-ichi Kawarabayashi,
  and Stefanie Jegelka.
\newblock Representation learning on graphs with jumping knowledge networks.
\newblock In Jennifer Dy and Andreas Krause, editors, \emph{Proceedings of the
  35th International Conference on Machine Learning}, volume~80 of
  \emph{Proceedings of Machine Learning Research}, pages 5453--5462. PMLR,
  10--15 Jul 2018{\natexlab{b}}.
\newblock URL \url{http://proceedings.mlr.press/v80/xu18c.html}.

\bibitem[Yang et~al.(2016)Yang, Cohen, and Salakhudinov]{yang2016revisiting}
Zhilin Yang, William Cohen, and Ruslan Salakhudinov.
\newblock Revisiting semi-supervised learning with graph embeddings.
\newblock In \emph{International conference on machine learning}, pages 40--48.
  PMLR, 2016.

\bibitem[Ying et~al.(2018)Ying, You, Morris, Ren, Hamilton, and
  Leskovec]{ying2018hierarchical}
Zhitao Ying, Jiaxuan You, Christopher Morris, Xiang Ren, Will Hamilton, and
  Jure Leskovec.
\newblock Hierarchical graph representation learning with differentiable
  pooling.
\newblock In \emph{Advances in neural information processing systems}, pages
  4800--4810, 2018.

\bibitem[Zeng et~al.(2019)Zeng, Zhou, Srivastava, Kannan, and
  Prasanna]{zeng2019graphsaint}
Hanqing Zeng, Hongkuan Zhou, Ajitesh Srivastava, Rajgopal Kannan, and Viktor
  Prasanna.
\newblock Graphsaint: Graph sampling based inductive learning method.
\newblock In \emph{International Conference on Learning Representations}, 2019.

\bibitem[Zhang et~al.(2018)Zhang, Cui, Neumann, and Chen]{zhang2018end}
Muhan Zhang, Zhicheng Cui, Marion Neumann, and Yixin Chen.
\newblock An end-to-end deep learning architecture for graph classification.
\newblock \emph{Proceedings of the AAAI Conference on Artificial Intelligence},
  32\penalty0 (1), Apr. 2018.
\newblock URL \url{https://ojs.aaai.org/index.php/AAAI/article/view/11782}.

\bibitem[Zhao and Akoglu(2020)]{Zhao2020PairNorm}
Lingxiao Zhao and Leman Akoglu.
\newblock Pairnorm: Tackling oversmoothing in {GNN}s.
\newblock In \emph{International Conference on Learning Representations}, 2020.
\newblock URL \url{https://openreview.net/forum?id=rkecl1rtwB}.

\bibitem[Zhou et~al.(2020{\natexlab{a}})Zhou, Cui, Hu, Zhang, Yang, Liu, Wang,
  Li, and Sun]{zhou2020graph}
Jie Zhou, Ganqu Cui, Shengding Hu, Zhengyan Zhang, Cheng Yang, Zhiyuan Liu,
  Lifeng Wang, Changcheng Li, and Maosong Sun.
\newblock Graph neural networks: A review of methods and applications.
\newblock \emph{AI Open}, 1:\penalty0 57--81, 2020{\natexlab{a}}.

\bibitem[Zhou et~al.(2020{\natexlab{b}})Zhou, Huang, Li, Zha, Chen, and
  Hu]{zhou-et-al-neurips-2020}
Kaixiong Zhou, Xiao Huang, Yuening Li, Daochen Zha, Rui Chen, and Xia Hu.
\newblock Towards deeper graph neural networks with differentiable group
  normalization.
\newblock In H.~Larochelle, M.~Ranzato, R.~Hadsell, M.~F. Balcan, and H.~Lin,
  editors, \emph{Advances in Neural Information Processing Systems}, volume~33,
  pages 4917--4928. Curran Associates, Inc., 2020{\natexlab{b}}.
\newblock URL
  \url{https://proceedings.neurips.cc/paper/2020/file/33dd6dba1d56e826aac1cbf23cdcca87-Paper.pdf}.

\end{thebibliography}
\bibliographystyle{plainnat}
\newpage
\appendix
\onecolumn
\section{Appendix}
\label{sec:appendix}

\subsection{Deep Supervision for Node Classification}
\label{sec:ds_for_node_classification}

In Section~\ref{sec:ds_gnns} we extended graph neural networks with deep supervision focused on the graph property prediction setting. Here, we explain how deep supervision can be applied for node property prediction with a focus on classification tasks.

We are given a graph represented as the tuple $G=(V, E)$ where $V$ is the set of nodes and $E$ the set of edges. The graph has $|V| = N$ nodes. We assume that each node $v \in V$ is also associated with an attribute vector $\mathbf{x}_v \in \R^d$ and let $\mathbf{X} \in \R^{N \times d}$ represent the attribute vectors for all nodes in the graph. 
A subset of $M$ nodes, $V_l \subset V$, has known labels. Each label represents one of $C$ classes using a one-hot vector representation such that $\mathbf{Y} \in \R^{M\times C}$. The node property prediction task is to learn a function $f: V \rightarrow Y$ that maps node representations to class probabilities.  

Consider the case of a $2$-layer GNN with GAT~\citep{velickovic2018graph} layers and one attention head.  
The node representations output by each of the $2$ GAT layers are given by,
\begin{align}
   \mbH^{(1)} = \relu(\mbOmega \mbX \mbW^{(1)}), \quad    \\
   \mbH^{(2)} = \relu(\mbOmega \mbH^{(1)} \mbW^{(2)}), \quad
\end{align}
where the $\relu$ activations are element-wise, $\mathbf{\Omega}$ are the attention weights given by Equation~\ref{eq:gat}, and $\mathbf{W}^{(i)}$ are trainable layer weights.

Let each GAT layer be followed by a linear layer with $\softmax$ activation calculating class probabilities for all nodes in the graph such that,
\begin{equation}
    \mathbf{Z}^{(l)} = \softmax(\mathbf{H}^{(l)}\mathbf{\widehat{W}}^{(l)}), \quad l=1,2,
    \label{eq:linear_gat_node_example}
\end{equation}
where $\mathbf{Z}^{(l)}$ are the class probabilities for all nodes as predicted by the $l$th layer, and $\mathbf{\widehat{W}}^{(l)}$ are the layer's trainable weights.

Now we can compute layer-dependent losses as:
\begin{equation}
    \cL_N^{(l)} = -\sum_{v \in V_l} \sum_{c=0}^{C-1} \mbY_{v, c}log(\mathbf{Z}^{(l)}_{v, c}), \quad l=1,2.
    \label{eq:node_cross_entropy}
\end{equation}

For a standard GNN, in order to estimate the weights $\{ \mathbf{W}^{(1)}, \mathbf{W}^{(2)}, \mathbf{\widehat{W}}^{(2)}  \}$, we optimize the cross-entropy loss calculated over the set of nodes with known labels only using $\cL_N^{(2)}$. 

Deep supervision adds a linear layer corresponding to each GAT layer in the model such that, in our example, the model makes two predictions for each node, $\mathbf{Z}^{(1)}$ and $\mathbf{Z}^{(2)}$. We now estimate the weights \{$ \mathbf{{W}}^{(1)}, \mathbf{W}^{(2)}, \mathbf{\widehat{W}}^{(1)}, \mathbf{\widehat{W}}^{(2)}$\}, and optimize the mean loss that for our example is given by,
\begin{equation}
    \cL_N = \frac{1}{2}\sum_{k=1}^{2} \cL^{(k)}_N.
    \label{eq:ds_node_loss}
\end{equation}

\subsection{Datasets}
\label{sec:dataset_stats}

\begin{table}[ht]
    \caption{Dataset statistics we used for the empirical evaluation. The number of nodes for \esol, \lipo, and \enzymes is the average of the number of nodes in all the graphs in each dataset. A value of `-' for train/val/test for the graph datasets indicates that 10-fold cross validation was used.}
    \vskip 0.15in
    \centering
    \begin{tabular}{lccccc}
        \toprule
        Name & Graphs & Nodes & Classes & Node features & \# train/val/test \\
        \midrule
        \enzymes & 600 &  33 (avg) & 6 & 18 & -\\
        \esol & 1144 & 13 (avg) & Regr. & 74 & -\\
        \lipo & 4200 & 27 (avg) & Regr. & 74 & -\\
        \cora & 1 & 2708 &7 & 1433 & 140/500/1000\\
        \citeseer & 1 &  3327 & 6 & 3703 & 120/500/1000\\
        \pubmed & 1 & 19717 &3 & 500 & 60/500/1000\\
        \bottomrule
    \end{tabular}
    \label{tab:datasets}
    \vskip -0.1in
\end{table}
Table~\ref{tab:datasets} gives detailed information about the datasets we used for the empirical evaluation of the different architectures. 

\cora, \citeseer, and \pubmed are citation networks where the goal is to predict the subject of a paper. Edges represent citation relationships. We treat these graphs as undirected as it is common in the GNN literature. The datasets have known train/val/test splits from \citet{yang2016revisiting}. Training sets are small with the number of labeled nodes equal to $140$ ($20$ for each of $7$ classes), $120$ ($20$ for each of $6$ classes), and $60$ ($20$ for each of $3$ classes) for \cora, \citeseer, and \pubmed respectively.

\enzymes is a graph classification dataset where the goal is to predict enzyme class as it relates to the reactions catalyzed. \esol is a regression dataset where the goal is to predict molecular solubility. Lastly, \lipo is a regression dataset where the goal is to predict the octanol/water distribution coefficient for a large number of compounds.

\subsection{Additional Experimental Results}
\label{sec:additional_experimental_results}

\subsubsection{Deep Model Performance}
\label{sec:node_deep_model_performance}

In Sections \ref{sec:results_node_classification} and \ref{sec:missing_features}, we noted that for deep models with $25$ layers, DSGNNs demonstrate better resilience to the over-smoothing problem. Our conclusion holds for both the normal and missing feature settings as can be seen in Figures \ref{fig:deep_gnn_graph_performance}, \ref{fig:deep_gnn_node_performance} and \ref{fig:deep_gnn_node_mv_performance}. 

Tables \ref{tab:results_summary_node_datasets_25} and \ref{tab:results_summary_node_datasets_25_mv} focus on the node classification performance of models with $25$ layers. In the normal setting (Table \ref{tab:results_summary_node_datasets_25}), DSGNN outperforms the others on \cora and \pubmed by $0.9\%$ and $1.3\%$ respectively whereas it is second best to JKNet on \citeseer trailing by $1.2\%$. In the missing feature setting (Table \ref{tab:results_summary_node_datasets_25_mv}), DSGNN outperforms the second best model on all three datasets by $1.7\%$, $2.3\%$, and $3.2\%$ for \cora, \citeseer, and \pubmed respectively. In the missing features setting, DSGNN outperforms a standard GNN with PairNorm by $4.4\%$, $2.3\%$, and $3.5\%$ for \cora, \citeseer, and \pubmed respectively. This evidence supports our conclusion that enhancing GNNs with deep supervision as opposed to PairNorm or jump connections is a more suitable solution to the over-smoothing problem for deep GNNs.
\begin{table}[ht]
    \caption{Node classification performance comparison between the standard GNN, Jumping Knowledge Network (JKNet) and the Deeply-Supervised GNN (DSGNN) architectures with and without PairNorm (PN). All models have $25$ layers. Results shown are mean accuracy and standard deviation on the test sets for $20$ repeats for fixed train/val/test splits. For each dataset, we use bold font to indicate the best performing model and underline the second best.}
    \vskip 0.15in
    \begin{center}
    \begin{tabular}{l  l  l  l}
        \toprule
        Model &  \cora & \citeseer & \pubmed \\ 
              &  \multicolumn{3}{c}{Accuracy $\uparrow$} \\
        \midrule
        GNN        &  74.3 $\pm$ 1.3              & 45.8 $\pm$ 3.3                   &  75.9	$\pm$ 0.9 \\
        JKNet      &  \underline{76.0 $\pm$ 1.5}  & \textbf{63.3	$\pm$ 2.3}       &  \underline{76.6 $\pm$ 0.8} \\ 
        DSGNN      &  \textbf{76.9 $\pm$ 0.8}     & \underline{62.1  $\pm$ 1.0}      &  \textbf{77.9	$\pm$ 0.5} \\ 
        GNN-PN     &  71.5 $\pm$ 0.9              & 57.0	$\pm$ 1.4                &  75.6	$\pm$ 0.7 \\
        DSGNN-PN   &  72.2 $\pm$ 0.6              & 51.3  $\pm$ 0.9                  &  74.2	$\pm$ 1.0 \\ 
        \bottomrule
    \end{tabular}
    \label{tab:results_summary_node_datasets_25}
    \end{center}
    \vskip -0.1in
\end{table}
\begin{table}[ht]
    \caption{\textbf{Missing feature setting} node classification performance comparison between the standard GNN, Jumping Knowledge Network (JKNet) and the Deeply-Supervised GNN (DSGNN) architectures with and without PairNorm (PN). All models have $25$ layers. Results shown are mean accuracy and standard deviation on the test sets for $20$ repeats for fixed train/val/test splits. For each dataset, we use bold font to indicate the best performing model and underline the second best.}
    \vskip 0.15in
    \begin{center}
    \centering
    \begin{tabular}{l  l  l  l}
        \toprule
        Model &  Cora & Citeseer & Pubmed \\ 
              &  \multicolumn{3}{c}{Accuracy $\uparrow$} \\
        \midrule
        GNN        & 72.6 $\pm$ 1.4              &   40.6 $\pm$ 2.9              &  75.7	$\pm$ 1.0\\
        JKNet      & \underline{73.9 $\pm$ 1.5}  &   50.3 $\pm$ 2.4              &  \underline{76.7	$\pm$ 0.9} \\ 
        DSGNN      & \textbf{75.6 $\pm$ 0.9}     &   \textbf{56.3 $\pm$ 2.2}     &  \textbf{77.9	$\pm$ 0.4} \\ 
        GNN-PN     & 71.2 $\pm$ 0.9              &   \underline{54.0 $\pm$ 1.6}  &  74.4	$\pm$ 1.1 \\
        DSGNN-PN   & 72.5 $\pm$ 0.8              &   50.3 $\pm$ 1.4              &  74.7	$\pm$ 0.9 \\ 
        \bottomrule
    \end{tabular}
    \label{tab:results_summary_node_datasets_25_mv}
    \end{center}
    \vskip -0.1in
\end{table}

\subsubsection{Analysis of Learned Representations}
\label{sec:learned-representations}
We provide additional evidence that DSGNNs learn more discriminate node representations for all hidden graph convolutional layers leading to performance benefits outlined above. We focus on the node classification domain. We adopt the metrics suggested by \citet{Zhao2020PairNorm} for measuring how discriminate node representations and node features are. 

Given a graph with $N$ nodes, let $\mathbf{H}^{(i)} \in \R^{N\times d}$ hold the $d$-dimensional node representations output by the $i$-th GAT layer. The row difference (row-diff) measures the average pairwise distance between the node representations (rows of $\mathbf{H}^{(i)}$). The column difference (col-diff) measures the average pairwise distance between the $L_1$-normalized columns of $\mathbf{H}^{(i)}$. The former measures node-wise over-smoothing and the latter feature-wise over-smoothing~\citep{Zhao2020PairNorm}.

\begin{figure}[ht]
	\begin{center}
	    \begin{tabular}{c}
        	\includegraphics[width=16.5cm]{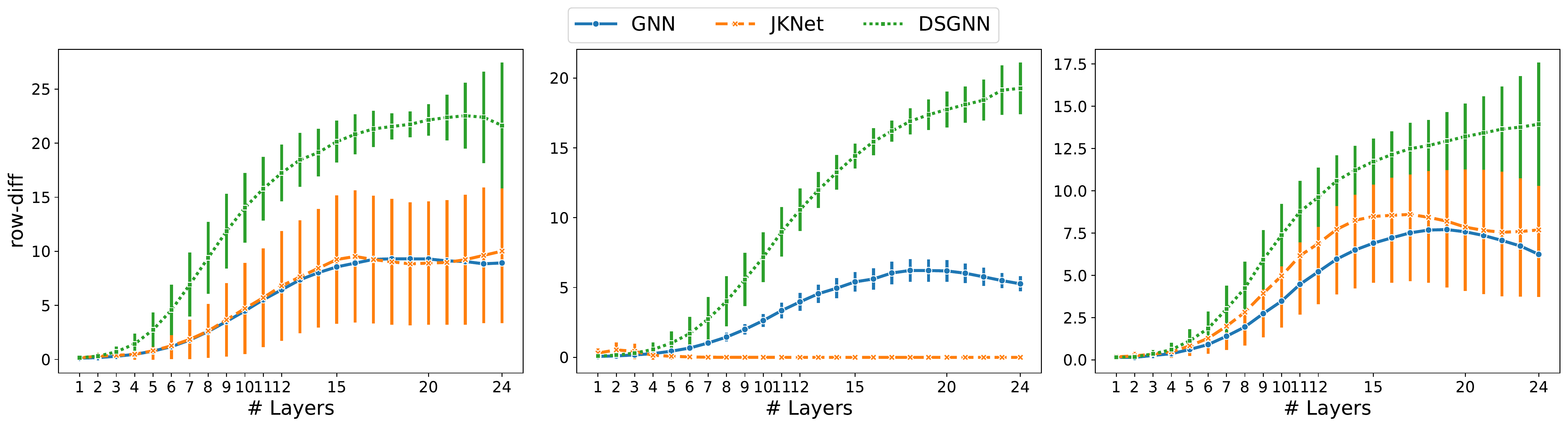} \\
        \end{tabular}
    \caption{Row difference metric from \citet{Zhao2020PairNorm} calculated for the output of each GAT layer in a $25$-layer model. Metric shown for \cora (left), \citeseer (middle), and \pubmed (right) datasets.}
	\label{fig:row_diff_25_layers}
	\end{center}
\end{figure}

\begin{figure}[ht]
	\begin{center}
	    \begin{tabular}{c}
        	\includegraphics[width=16.5cm]{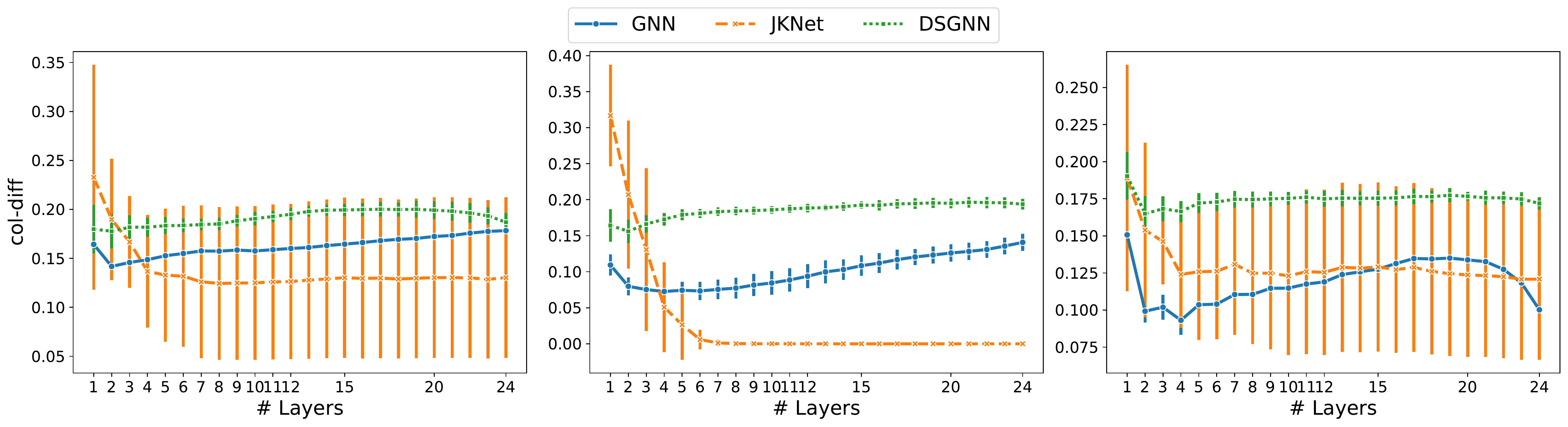} \\
        \end{tabular}
    \caption{Column difference metric from \citet{Zhao2020PairNorm} calculated for the output of each GAT layer in a $25$-layer model. Metric shown for \cora (left), \citeseer (middle), and \pubmed (right) datasets.}
	\label{fig:col_diff_25_layers}
	\end{center}
\end{figure}

We consider the row-diff and col-diff metrics for the deepest models we trained, those with $25$ layers of which $24$ are GAT. We calculate the two metrics for the node representations output by each of the GAT layers. Figures \ref{fig:row_diff_25_layers} and \ref{fig:col_diff_25_layers} show plots of the row-diff and col-diff metrics respectively. We note that for all datasets, DSGNN node representations are the most separable for the majority of layers. For all models, row-diff plateaus after the first few layers. We interpret this as a point of convergence for the learnt node representations such that adding more layers can only harm the model's performance as indicated in Figure \ref{fig:deep_gnn_node_performance}. We further demonstrate this point by visualising the node embeddings for \cora.

\begin{figure}[ht]
	\begin{center}
	    \begin{tabular}{c}
        	\includegraphics[width=16.5cm]{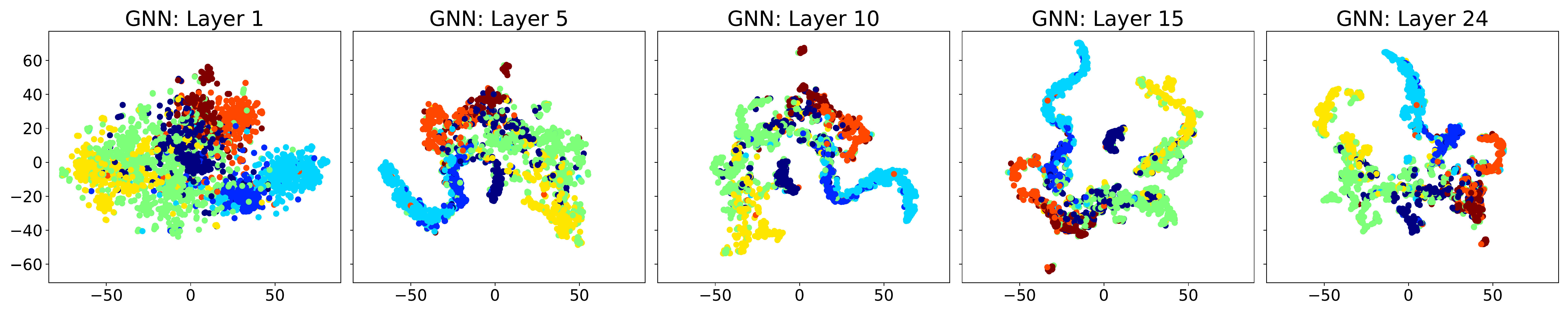} \\
            \includegraphics[width=16.5cm]{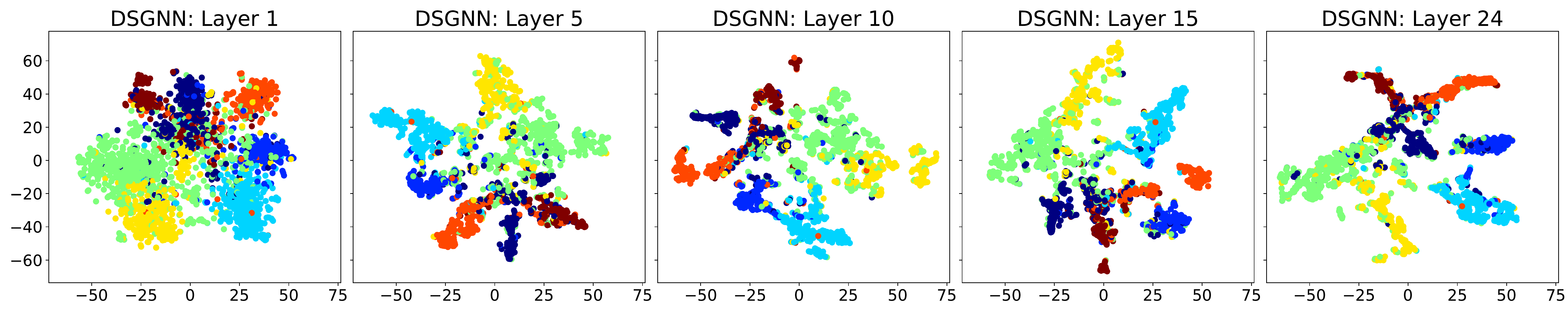} \\
        	\includegraphics[width=16.5cm]{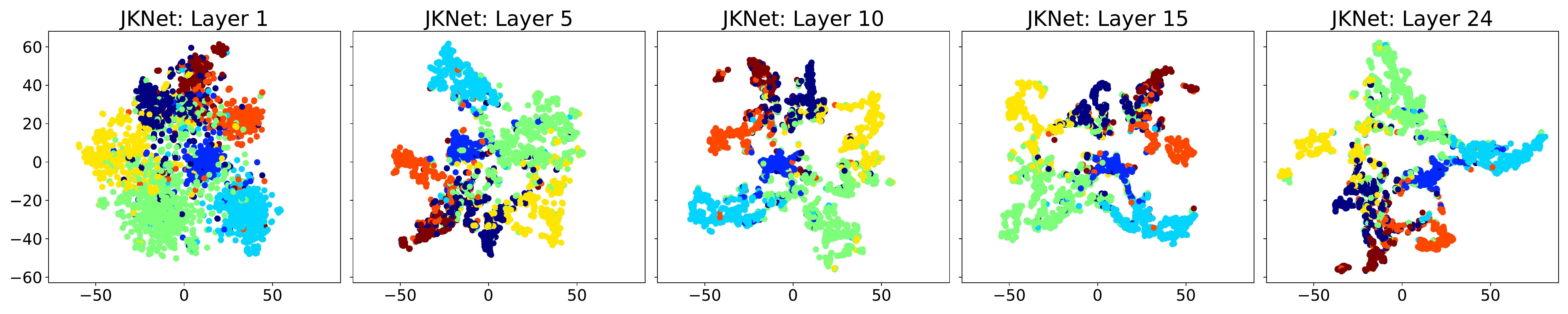} \\            
        \end{tabular}
    \caption{\cora node embeddings for the standard GNN (top), DSGNN (middle) and JKNet (bottom) models each with $25$ layers. We show the node embeddings output from layers $1$ (the first GAT layer), $5$, $10$, $15$, and $24$ (the last GAT layer in all models). The colors indicate node class.}
	\label{fig:node_embeddings_cora}
	\end{center}
\end{figure}

Figure \ref{fig:node_embeddings_cora} shows a visual representation of the learnt node embeddings for a subset of the GAT layers in the trained $25$-layer models. We used t-SNE~\citep{JMLR:v9:vandermaaten08a} to project the $64$-dimensional node features to $2$ dimensions. We can see that all architectures learn clusters of nodes with similar labels. However, for the standard GNN and JKNet, these clusters remain the same for the $10$-th layer and above. On the other hand, DSGNN continues to adapt the clusters for all layers as we would expect given the effect of the companion losses associated with each GAT layer.

\end{document}